\documentclass[runningheads]{llncs}
\usepackage{graphicx}
\usepackage{comment}
\usepackage{amsmath,amssymb}
\usepackage{url,multirow,tabulary}
\usepackage{color}
\usepackage{subcaption}
\usepackage[cmintegrals]{newtxmath}
\usepackage[utf8]{inputenc}
\usepackage{newunicodechar}
\usepackage{makecell}
\usepackage{booktabs}
\usepackage{array}
\usepackage[table]{xcolor}
\usepackage{mathtools}
\usepackage[colorlinks,bookmarks=false]{hyperref}
\usepackage{float}
\usepackage[width=122mm,left=12mm,paperwidth=146mm,height=193mm,top=12mm,paperheight=217mm]{geometry}
\usepackage{mathrsfs}

\newcommand{\etal}{\textit{et al.}}

\newcolumntype{L}[1]{>{\raggedright\arraybackslash}p{#1}}
\newcolumntype{C}[1]{>{\centering\arraybackslash}p{#1}}
\newcolumntype{R}[1]{>{\raggedleft\arraybackslash}p{#1}}

\begin{document}
\pagestyle{headings}\mainmatter\def\ECCVSubNumber{1}

\title{We Learn Better Road Pothole Detection:\\from Attention Aggregation to\\ Adversarial Domain Adaptation}

\titlerunning{We Learn Better Road Pothole Detection}
\authorrunning{R. Fan, H. Wang, M. J. Bocus, and M. Liu}

\author{Rui Fan\inst{1}$^\star$ \and
    Hengli Wang\inst{2}\thanks{These authors contributed equally to this work and are therefore joint first authors.} \and
    Mohammud J. Bocus\inst{3} \and Ming Liu\inst{2}}
\institute{UC San Diego \\
    \email{rui.fan@ieee.org} \and
    HKUST Robotics Institute \\
    \email{\{hwangdf, eelium\}@ust.hk}
\and
    University of Bristol\\
    \email{junaid.bocus@bristol.ac.uk}
}

\maketitle

\begin{abstract}
Manual visual inspection performed by certified inspectors is still the main form of road pothole detection. This process is, however, not only tedious, time-consuming and costly, but also dangerous for the inspectors. Furthermore, the road pothole detection results are always subjective, because they depend entirely on the individual experience. Our recently introduced disparity (or inverse depth) transformation algorithm allows better discrimination between damaged and undamaged road areas, and it can be easily deployed to any semantic segmentation network for better road pothole detection results. To boost the performance, we propose a novel attention aggregation (AA) framework, which takes the advantages of different types of attention modules. In addition, we develop an effective training set augmentation technique based on adversarial domain adaptation, where the synthetic road RGB images and transformed road disparity (or inverse depth) images are generated to enhance the training of semantic segmentation networks. The experimental results demonstrate that, firstly, the transformed disparity (or inverse depth) images become more informative; secondly, AA-UNet and AA-RTFNet, our best performing implementations, respectively outperform all other state-of-the-art single-modal and data-fusion networks for road pothole detection; and finally, the training set augmentation technique based on adversarial domain adaptation not only improves the accuracy of the state-of-the-art semantic segmentation networks, but also accelerates their convergence.

\end{abstract}

\begin{center}
    \textbf{Source Code and Dataset:}\\
    \url{sites.google.com/view/pothole-600}
\end{center}

\section{Introduction}
\label{sec.introduction}
Potholes are small concave depressions on the road surface \cite{mathavan2015review}. They arise due to a number of environmental factors, such as water permeating into the ground under the asphalt road surface \cite{fan2019pothole}. The affected road areas are further deteriorated due to the vibration of tires, making the road surface impracticable for driving. Furthermore, vehicular traffic can cause the subsurface materials to move, and this generates a weak spot under the street. With time, the road damage worsens due to the frequent movement of vehicles over the surface and this causes new road potholes to emerge. \cite{koch2015review}.

Road pothole is not just an inconvenience, but also poses a safety risk, because it can severely affect vehicle condition, driving comfort, and traffic safety \cite{fan2019pothole}. It was reported in 2015 that Danielle Rowe, an Olympic gold medalist as well as three-time world champion, had eight fractured ribs resulting in a punctured lung, after hitting a pothole during a race \cite{pothole_news_dani_king}. Therefore, it is crucial and necessary to regularly inspect road potholes and repair them in time.

Currently, manual visual inspection performed by certified inspectors is still the main form of road pothole detection \cite{fan2019road}. However, this process is not only time-consuming, exhausting and expensive, but also hazardous for the inspectors \cite{koch2015review}. For example, the city of San Diego repairs more than 30K potholes per year using hot patches compound and bagged asphalt, and they have been requesting residents to report potholes so as to relieve the burden on the local road maintenance department \cite{sandiego_pothole}. Elsewhere, the UK government is set to pledge billions of pounds for filling potholes across the country \cite{bbc_pothole_repairing}. Additionally, the pothole detection results are always subjective, as the decisions depend entirely on the inspector's experience and judgment \cite{fan2019real}. Hence, there has been a strong demand for automated road condition assessment systems, which can not only acquire 2D/3D road data, but also detect and predict road potholes accurately, robustly and objectively \cite{leo2018deep}.

Specifically, automated road pothole detection has been considered as more than an infrastructure maintenance problem in recent years, as many self-driving car companies have included road pothole detection into their autonomous car perception modules. For instance, Jaguar Land Rover announced their recent research achievements on road pothole detection/prediction \cite{land_rover_pothole}, where the vehicles can not only gather the location and severity data of the road potholes, but also send driver warnings to slow down the car. Ford also claimed that they were experimenting with data-driven technologies to warn drivers of the pothole locations \cite{ford_pothole}. Furthermore, during the Consumer Electronics Show (CES) 2020, Mobileye demonstrated their solutions\footnote{\url{s21.q4cdn.com/600692695/files/doc_presentations/2020/1/Mobileye-CES-2020-presentation.pdf}} for road pothole detection, which are based on machine vision and intelligence. With recent advances in image analysis and deep learning, especially for 3D vision data, depth/disparity image analysis and convolutional neural networks (CNNs) have become the mainstream techniques  for road pothole detection \cite{fan2019real}.

Given the 3D road data, image segmentation algorithms are typically performed to detect potholes. For example, Jahanshahi {\etal} \cite{jahanshahi2013unsupervised} employed Otsu's thresholding method \cite{otsu1979threshold} to segment depth images for road pothole detection. In \cite{fan2019pothole}, we proposed a disparity image transformation algorithm, which can better distinguish between damaged and undamaged road areas. The road potholes were then detected using a surface modeling approach. Subsequently, we minimized the computational complexity of our algorithm and successfully embedded it in a drone for real-time road inspection \cite{fan2019real}. Recently, the aforementioned algorithm was proved to have a numeric solution \cite{fan2019road}, which allows it to be easily deployed to any existing semantic segmentation networks for end-to-end road pothole detection.

In this paper, we first briefly introduce the disparity (or inverse depth, as disparity is in inverse proportion to depth) transformation (DT) algorithm proposed in \cite{fan2019road}. We then exploit the aggregation of different types of attention modules (AMs) so as to improve the semantic segmentation networks for better road pothole detection. Furthermore, we develop a novel adversarial domain adaptation framework for training set augmentation. Moreover, we publish our road pothole detection dataset, named \textit{Pothole-600}, at \url{sites.google.com/view/pothole-600} for research purposes. According to our experimental results presented in Section \ref{sec.exp}, training CNNs with augmented road data  yields better semantic segmentation results, where convergence is achieved with fewer iterations at the same time.

\section{Related Works}
\label{sec.related_works}

\subsection{Semantic Segmentation}
\label{sec.semantic_segmentation}
Fully convolutional network (FCN) \cite{long2015fully} was the first end-to-end single-modal CNN designed for semantic segmentation. Based on FCN, U-Net \cite{ronneberger2015u} adopts an encoder-decoder architecture. It also adds skip connections between the encoder and decoder to help smooth the gradient flow and restore the locations of objects. Additionally, PSPNet \cite{zhao2017pyramid}, DeepLabv3+ \cite{chen2018encoder} and DenseASPP \cite{yang2018denseaspp} leverage a pyramid pooling module to extract context information for better segmentation performance. Furthermore, GSCNN \cite{takikawa2019gated} employs a two-branch framework consisting of a shape branch and a regular branch, which can effectively improve the semantic predictions on the boundaries.

Different from the above-mentioned single-modal networks, many data-fusion networks have also been proposed to improve semantic segmentation accuracy by extracting and fusing the features from multi-modalities of visual information \cite{wang2020applying}, \cite{fan2020sne-roadseg}. For instance, FuseNet \cite{hazirbas2016fusenet} and depth-aware CNN \cite{wang2018depth} adopt the popular encoder-decoder architecture, but employ different operations to fuse the feature maps obtained from the RGB and depth branches. Moreover, RTFNet \cite{sun2019rtfnet} was developed to improve semantic segmentation performance by fusing the features extracted from RGB images and thermal images. It also adopts an encoder-decoder architecture and an element-wise addition fusion strategy.

\subsection{Attention Module}
\label{sec.attention_module}
Due to their simplicity and effectiveness, AMs have been widely used in various computer vision tasks. AMs typically learn the weight distribution (WD) of an input feature map and output an updated feature map based on the learned WD \cite{vaswani2017attention}. Specifically, Squeeze-and-Excitation Network (SENet) \cite{hu2018squeeze} employs a channel-wise AM to improve image classification accuracy. Furthermore, Wang {\etal} \cite{wang2018non} presented a non-local module to capture long-range dependencies for video classification. OCNet \cite{yuan2018ocnet} and DANet \cite{fu2019dual} proposed different self-attention modules that are capable of using contextual information for semantic segmentation. Moreover, CCNet \cite{huang2019ccnet} adopts a criss-cross AM to obtain dense contextual information in a more efficient way. Different from the aforementioned studies, we propose an attention aggregation (AA) framework that focuses on the combination of different AMs. Based on this idea, our proposed AA-UNet and AA-RTFNet can take advantage of different AMs and yield accurate results for road pothole detection.

\subsection{Adversarial Domain Adaptation}
Since the concept of ``generative adversarial network (GAN)'' \cite{goodfellow2014generative} was first introduced in 2014, great efforts have been made in this research area to improve the existing computer vision algorithms. The recipe for their success is the use of an adversarial loss, {which makes the generated synthetic images become indistinguishable from the real images when minimized \cite{zhu2017unpaired}.}

Recent image-to-image translation approaches typically utilize a dataset,  which contains paired source and target images, to learn a parametric translation using CNNs. One of the most well-known work is the ``pix2pix'' framework \cite{isola2017image} proposed by Isola {\etal}, which employs a conditional GAN to learn the mapping from source images to target images.

In addition to the paired image-to-image translation approaches mentioned above, many unsupervised approaches have also been proposed  in recent years to tackle unpaired image-to-image translation problem, where the primary goal is to learn a mapping $G: \mathcal{S}\rightarrow\mathcal{T}$ from source domain $\mathcal{S}$ to target domain $\mathcal{T}$, so that the distribution of images from $G(\mathcal{S})$ is indistinguishable from the distribution $\mathcal{T}$. CycleGAN \cite{zhu2017unpaired} is a representative work handling unpaired image-to-image translation, where an inverse mapping $F:\mathcal{T}\rightarrow\mathcal{S}$ and a cycle-consistency loss (aiming at forcing $F(G(\mathcal{S}))\approxeq \mathcal{S}$) were coupled with $G: \mathcal{S}\rightarrow\mathcal{T}$. Our proposed training set augmentation technique is developed based on CycleGAN \cite{zhu2017unpaired}, but it performs paired image-to-image translation.

\section{Disparity (or Inverse Depth) Transformation}
\label{sec.quasi_dense_inverse_perspective}

\begin{figure*}[t]
    \centering
    \includegraphics[width=0.70\textwidth]{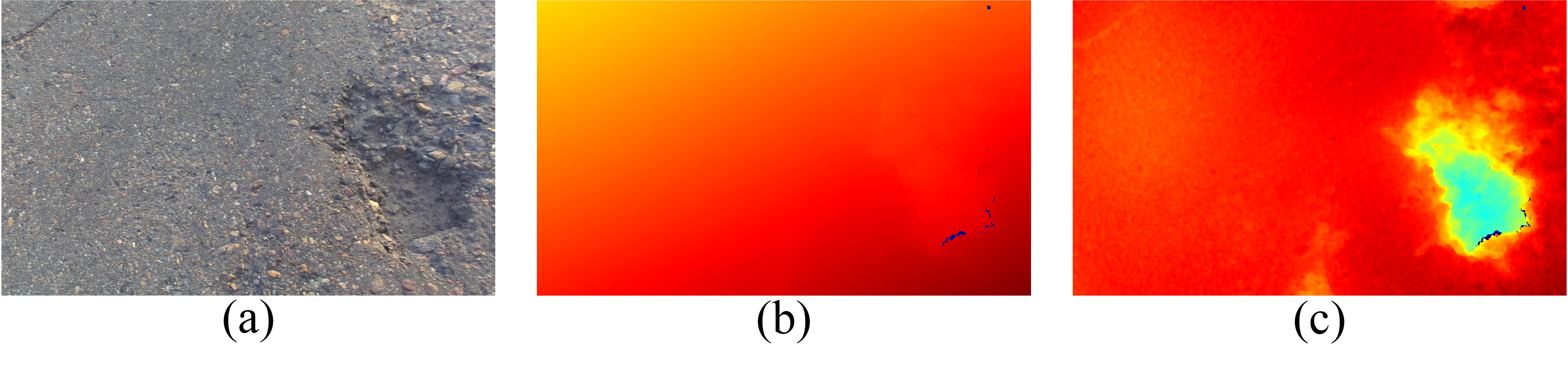}
    \caption{Disparity transformation: (a) RGB image; (b) disparity image produced by PT-SRP \cite{fan2018road}; and (c) transformed disparity image.}
    \label{fig.qipt}
\end{figure*}

DT aims at transforming a disparity or inverse depth image $\mathbf{G}$ into a quasi bird's eye view, whereby the pixels in the undamaged road areas possess similar values, while they differ significantly from those of the pothole pixels.

Since the concept of ``v-disparity domain'' was introduced in \cite{labayrade2003single}, disparity image analysis has become a common technique used for 3D driving scene understanding \cite{fan2019real}. The projections of the on-road disparity (or inverse depth) pixels in the v-disparity domain can be represented by a non-linear model as follows:
\begin{equation}
\tilde{\mathbf{q}}=\mathbf{M}\tilde{\mathbf{p}}={\varkappa}\begin{bmatrix*}[r]
-\sin\Phi & \cos\Phi & \kappa\\
0 & {1}/{\varkappa} & 0 \\
0 & 0 & {1}/{\varkappa}
\end{bmatrix*}\tilde{\mathbf{p}},
\end{equation}
where $\tilde{\mathbf{p}}=[u,v,1]^\top$ is the homogeneous coordinates of a pixel in the disparity (or inverse depth) image, and $\tilde{\mathbf{q}}=[g,v,1]^\top$ is the homogeneous coordinates of its projection in the v-disparity domain. $\Phi$ can be estimated via \cite{fan2019real}:
\begin{equation}
\underset{\Phi}{\arg\min}\  \mathbf{g}^\top\mathbf{g}-\mathbf{g}^\top\mathbf{T}(\Phi)\big(\mathbf{T}(\Phi)^\top\mathbf{T}(\Phi)\big)^{-1}\mathbf{T}(\Phi)^\top\mathbf{g},
\label{eq.energy}
\end{equation}
where $\mathbf{g}$ is a $k$-entry vector of disparity (or inverse depth) values, $\mathbf{1}_k$ is a $k$-entry vector of ones, $\mathbf{u}$ and $\mathbf{v}$ are two $k$-entry vectors storing the horizontal and vertical coordinates of the observed pixels, respectively, and $\mathbf{T}(\Phi)=[\mathbf{1}_k,\cos\Phi \mathbf{v}-\sin\Phi \mathbf{u}]$.  (\ref{eq.energy}) has a closed-form solution as follows \cite{fan2019road}:
\begin{equation}
\Phi=
\arctan\frac{\omega_4\omega_0-\omega_3\omega_1+q\sqrt{\Delta}}{\omega_3\omega_2+\omega_5\omega_1-\omega_5\omega_0-\omega_4\omega_2}
\ \ \text{s.t.} \ q\in\{-1,1\},
\label{eq.phi}
\end{equation}
where
\begin{equation}
\Delta=(\omega_4\omega_0-\omega_3\omega_1)^2+(\omega_3\omega_2-\omega_5\omega_0)^2-(\omega_4\omega_2-\omega_5\omega_1)^2.
\label{eq.delta}
\end{equation}
The expressions of $\omega_0$-$\omega_5$ are given in \cite{fan2019road}. $\kappa$ and $\varkappa$ can then be obtained using:
\begin{equation}
\mathbf{x}=	\varkappa\begin{bmatrix}
    \kappa \\
    1
    \end{bmatrix}=\big(\mathbf{T}(\Phi)^\top\mathbf{T}(\Phi)\big)^{-1}\mathbf{T}(\Phi)^\top\mathbf{g}.
\end{equation}
DT can therefore be realized using \cite{fan2019road}:
\begin{equation}
\mathbf{G}'(\mathbf{p})=\mathbf{G}(\mathbf{p})-\varkappa\big(\cos\Phi v - \sin\Phi u \big) - \varkappa\kappa + \Lambda,
\end{equation}
where $\Lambda$ is a constant used to ensure that the values in the transformed disparity (or depth inverse) image $\mathbf{G}'$ are non-negative.
An example of the transformed disparity (or inverse depth) image is shown in Fig. \ref{fig.qipt}, where it can be observed that the damaged road area becomes highly distinguishable. The effectiveness of DT on improving semantic segmentation is discussed in Section \ref{sec.performance_evaluation}.

\section{Attention Aggregation Framework}
\label{sec.attention_combination_framework}

\begin{figure*}[t]
    \centering
    \includegraphics[width=\textwidth]{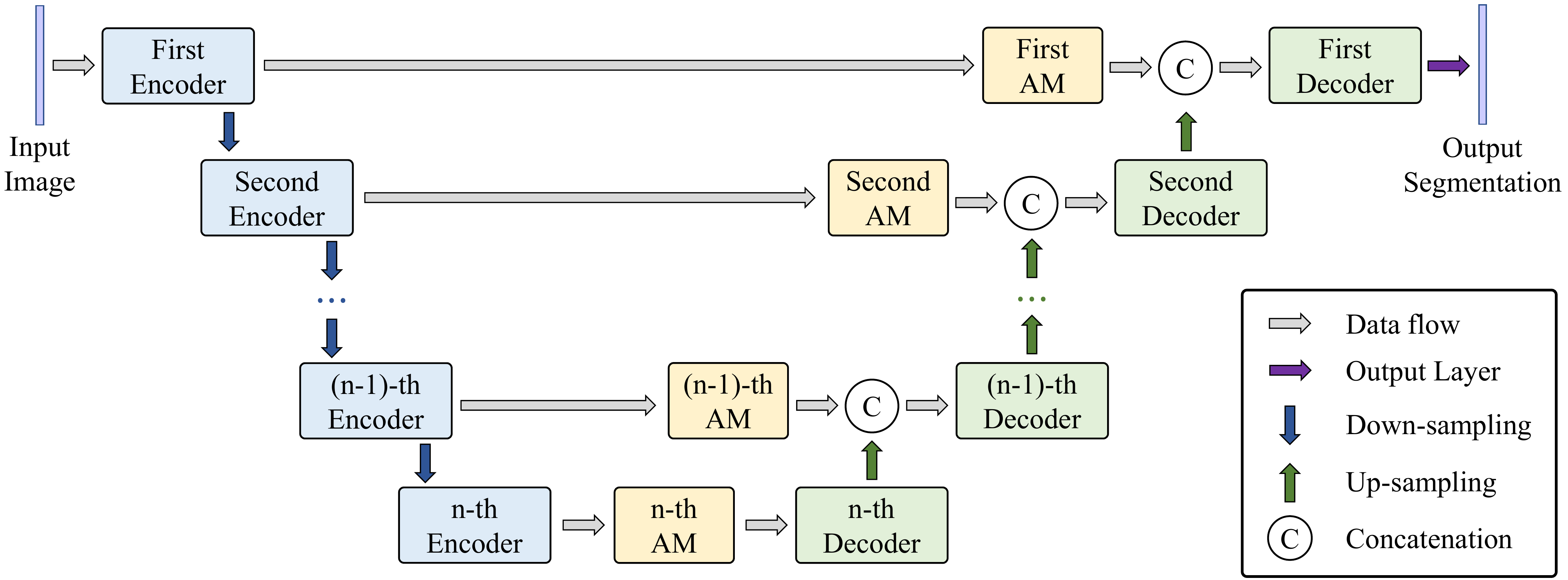}
    \caption{The architecture of the proposed attention aggregation framework for our AA-UNet and AA-RTFNet.}
    \label{fig.acnet}
\end{figure*}

The architecture of our proposed attention aggregation framework is illustrated in Fig. \ref{fig.acnet}. We add different AMs into the existing CNNs that adopt the popular encoder-decoder architecture. Firstly, U-Net \cite{ronneberger2015u} has demonstrated the effectiveness of employing skip connections, which concatenate the same-scale feature maps produced by the encoder and decoder. However, these two feature maps can present large difference because of the different numbers of transformations undergone, which can result in significant performance degradation. To alleviate this drawback, we add an AM for the encoder feature map before the concatenation in each skip connection, as shown in Fig. \ref{fig.acnet} (from the 1st to ($n-1$)-th AMs), where $n$ denotes the number of network levels. These AMs enable the encoder feature maps to focus on the potholes, which can shorten the gap between the same-scale feature maps produced by the encoder and decoder. This further improves pothole detection performance. Secondly, many studies \cite{fu2019dual,huang2019ccnet} have already demonstrated that adding an AM for a high-level feature map can significantly improve the overall performance. Therefore, we follow this paradigm and add an AM at the highest level, as shown in Fig. \ref{fig.acnet} ($n$-th AM).

\begin{figure*}[t]
    \centering
    \includegraphics[width=\textwidth]{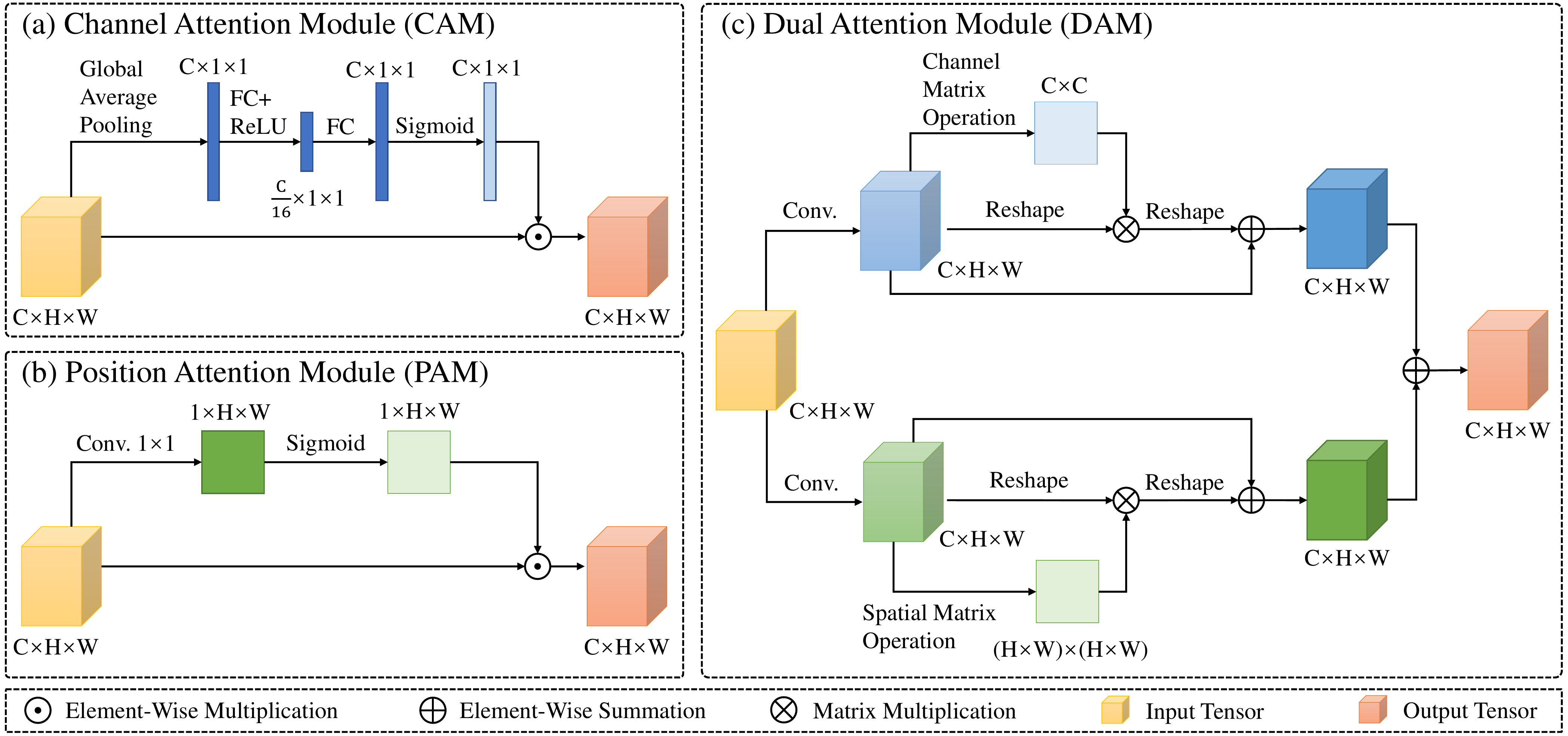}
    \caption{The illustrations of the three AMs used in our attention aggregation framework.}
    \label{fig.attention}
\end{figure*}

We use three AMs in our attention aggregation framework: 1) Channel Attention Module (CAM), 2) Position Attention Module (PAM) and 3) Dual Attention Module (DAM) \cite{fu2019dual}, as illustrated in Fig. \ref{fig.attention}. Similar to SENet \cite{hu2018squeeze}, our CAM is designed to assign each channel with a weight since some channels are more important. It first employs a global average pooling layer to squeeze spatial information, and then utilizes fully connected (FC) layers to generate the WD, which is finally combined with the input feature map by element-wise multiplication operation to generate the output feature map. Different from CAM, our PAM focuses on spatial information. It first generates the spatial WD and applies it on the input feature map to generate the output feature map. DAM \cite{fu2019dual} is composed of a channel attention submodule and a position attention submodule. Different from our CAM and PAM, these two submodules adopt the self-attention scheme to generate the WD, which can achieve better performance at the expense of a higher computational complexity. Since the memory consumed by DAM will grow significantly with the increase of feature map size, we only  use it at the highest level ($n$-th AM) so as to ensure computational efficiency.

To demonstrate the effectiveness of our framework, we employ it in a single-modal network (U-Net) and a data-fusion network (RTFNet), and dub them as AA-UNet and AA-RTFNet, respectively. The specific architectures (the selection of each AM) of our AA-UNet and AA-RTFNet are discussed  in Section \ref{sec.architecture_selection}.

\section{Adversarial Domain Adaptation for Training Set Augmentation}
\label{sec.gan_data_augmentation}

\begin{figure*}[t]
    \centering
    \includegraphics[width=\textwidth]{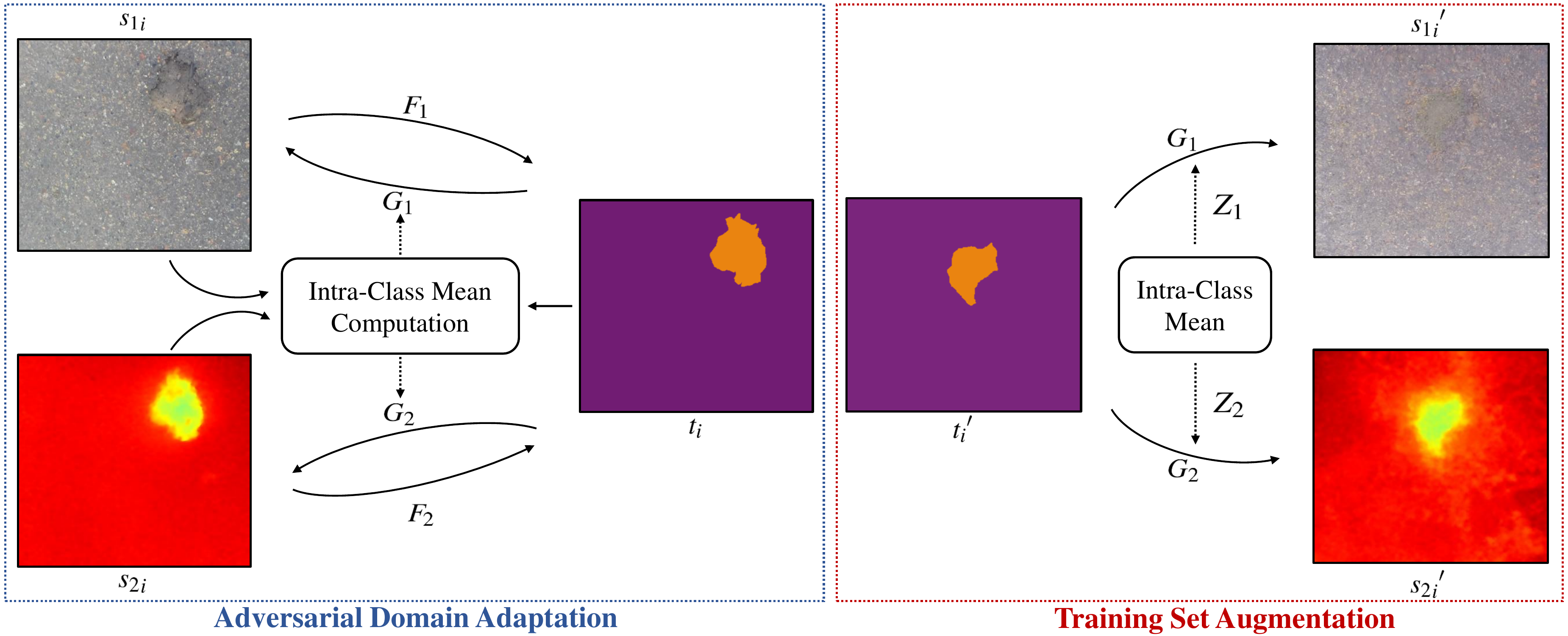}
    \caption{Adversarial domain adaptation for training set augmentation.}
    \label{fig.gan_block}
\end{figure*}

In this paper, adversarial domain adaptation is utilized to augment training set so that  the semantic segmentation networks can perform more robustly. Our proposed training set augmentation framework is illustrated in Fig. \ref{fig.gan_block}, where $F_1:\mathcal{S}_1\rightarrow\mathcal{T}$ translates RGB images ${s_1}_i\in\mathcal{S}_1$ to pothole detection ground truth $t_i\in\mathcal{T}$;
$G_1:\mathcal{T}\rightarrow\mathcal{S}_1$ translates
pothole detection ground truth $t_i\in\mathcal{T}$ back to RGB images ${s_1}_i\in\mathcal{S}_1$; $F_2:\mathcal{S}_2\rightarrow\mathcal{T}$ translates our transformed disparity images ${s_2}_i\in\mathcal{S}_2$ to pothole detection ground truth $t_i\in\mathcal{T}$;
and $G_2:\mathcal{T}\rightarrow\mathcal{S}_2$ translates
pothole detection ground truth $t_i\in\mathcal{T}$ back to our transformed disparity images ${s_2}_i\in\mathcal{S}_2$. The learning of $G_1$ and $G_2$ is guided by the intra-class means. Our full objective is:
\begin{equation}
\begin{split}
\mathcal{L}(G_1,G_2,F_1,F_2,D_{\mathcal{S}_1},D_{\mathcal{S}_2},D_{\mathcal{T}})
&=\mathcal{L}_\text{GAN}(G_1,D_{\mathcal{S}_1},\mathcal{T},{\mathcal{S}_1})+\mathcal{L}_\text{GAN}(F_1,D_{\mathcal{T}},{\mathcal{S}_1},\mathcal{T})\\
&+\mathcal{L}_\text{GAN}(G_2,D_{\mathcal{S}_2},\mathcal{T},{\mathcal{S}_2})+\mathcal{L}_\text{GAN}(F_2,D_{\mathcal{T}},{\mathcal{S}_2},\mathcal{T})\\
&+\mathcal{L}_\text{cyc}(G_1,F_1)+\mathcal{L}_\text{cyc}(G_2,F_2),
\end{split}
\end{equation}
where
\begin{equation}
\mathcal{L}_{GAN}(G,D_\mathcal{S},\mathcal{T},\mathcal{S})=
\mathbb{E}_{s\sim p_\text{data}(s)}[\log D_\mathcal{S}(s)]
+\mathbb{E}_{t\sim p_\text{data}(t)}[\log (1-D_\mathcal{S}(G(t)))],
\end{equation}
\begin{equation}
\mathcal{L}_{GAN}(F,D_\mathcal{T},\mathcal{S},\mathcal{T})=
\mathbb{E}_{t\sim p_\text{data}(t)}[\log D_\mathcal{T}(t)]
+\mathbb{E}_{s\sim p_\text{data}(s)}[\log (1-D_\mathcal{T}(F(s)))],
\end{equation}
\begin{equation}
\mathcal{L}_{cyc}(G,F)=
\mathbb{E}_{s\sim p_\text{data}(s)}[G(F(s))-s]
+\mathbb{E}_{t\sim p_\text{data}(t)}[F(G(t))-t],
\end{equation}
$D_\mathcal{S}$ and $D_\mathcal{T}$ are two adversarial discriminators: $D_\mathcal{S}$ aims to distinguish between images $\{s\}$ and the translated images $\{G(t)\}$, while $D_\mathcal{T}$ aims to distinguish between images $\{t\}$ and the translated images $\{F(s)\}$; $s\sim p_\text{data}(s)$ and $t\sim p_\text{data}(t)$ denote the data distributions of the source and target domains, respectively.

\begin{figure*}[t]
    \centering
    \includegraphics[width=\textwidth]{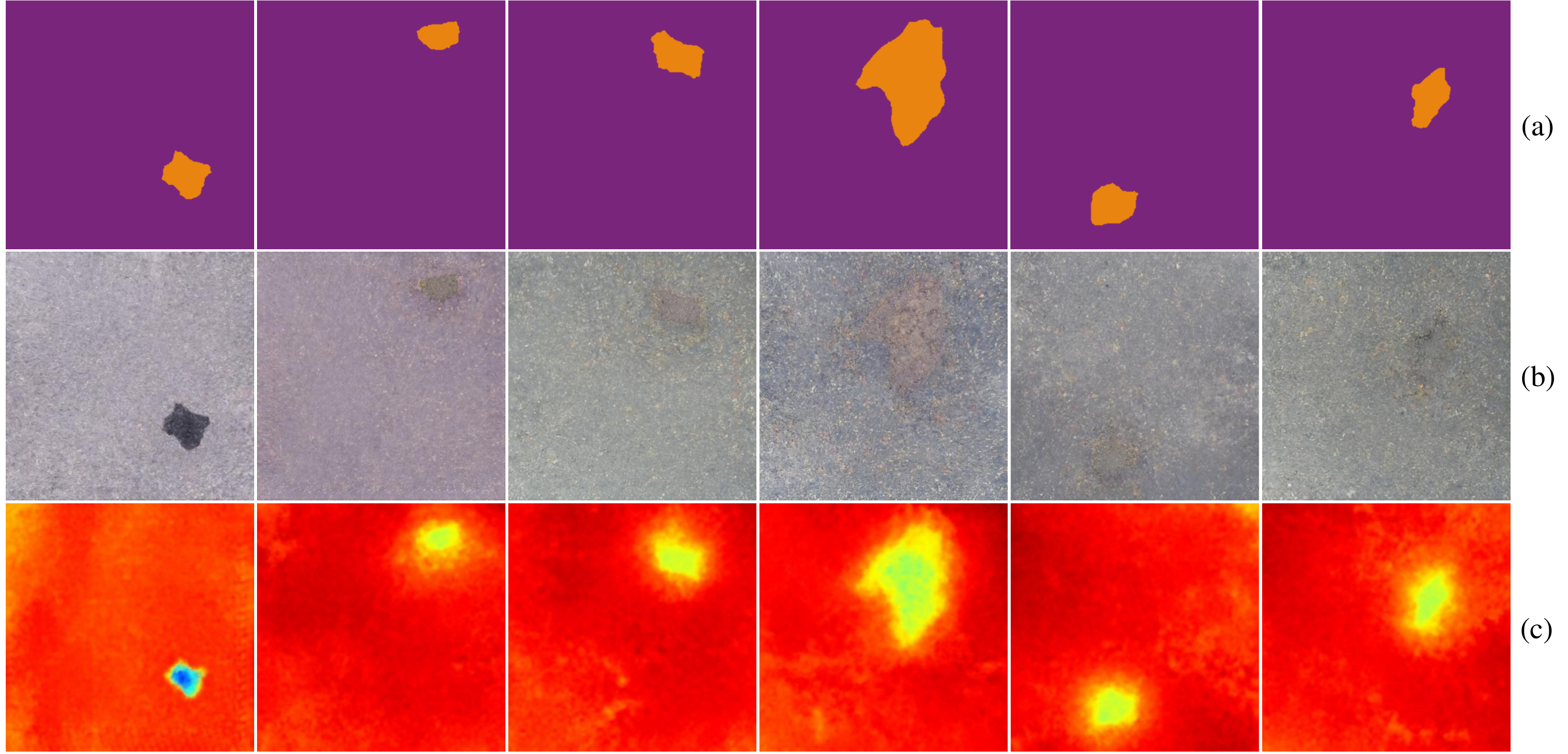}
    \caption{Examples of training set augmentation results: (a) randomly created pothole detection ground truth; (b) generated RGB images; and (c) generated transformed disparity images.}
    \label{fig.gan_results}
\end{figure*}

With well-learned mapping functions $G_1$ and $G_2$, we can generate an infinite number of synthetic RGB images ${{s_1}_i}'\in\mathcal{S}_1'$ and their corresponding synthetic transformed disparity images ${{s_2}_i}'\in\mathcal{S}_2'$ from a randomly generated pothole detection ground truth $t_i'\in\mathcal{T}'$. In order to expand the distributions of the two domains ${s_1}'\sim p_\text{data}({s_1}')$ and ${s_2}'\sim p_\text{data}({s_2}')$, we add random Gaussian noises $Z_1$ and $Z_2$ into $G_1$ and $G_2$ when generating ${{s_1}_i}'$ and ${{s_2}_i}'$, as shown in Fig. \ref{fig.gan_block}. Some examples in the augmented training set are shown in Fig. \ref{fig.gan_results}. The benefits of our proposed training set augmentation technique for semantic segmentation are discussed in Section \ref{sec.performance_evaluation}.

\section{Experiments}
\label{sec.exp}

\subsection{Datasets}
\label{sec.datasets}

\subsubsection{Pothole-600}
In our experiments, we utilized a stereo camera to capture stereo road images. These images are then split into a training set, a validation set and a testing set, which contains 240, 180 and 180 pairs of RGB images and transformed disparity images, respectively.

\subsubsection{Augmented Training Set}
We use adversarial domain adaptation to produce an augmented training set, which contains 2,400 pairs of RGB images and transformed disparity images. The performance comparison between using the original training set and using the augmented training set is presented in Section \ref{sec.performance_evaluation}.

\subsection{Experimental Setup}
\label{sec.setup}
In our experiments, we first select the architecture of our AA-UNet and AA-RTFNet, as presented in Section \ref{sec.architecture_selection}. Then, we compare our AA-UNet and AA-RTFNet with eight state-of-the-art (SoA) CNNs  (five single-modal ones and three data-fusion ones) for road pothole detection. Each single-modal CNN is trained using RGB images (RGB) and transformed disparity images (T-Disp), respectively; while each data-fusion CNN is trained using RGB and transformed disparity images (RGB+T-Disp). Furthermore, we also select different numbers of RGB images and transformed disparity images from our augmented training set to train the CNNs.  The experimental results are presented in Section \ref{sec.performance_evaluation}.

To quantify the performance of these CNNs, we adopt the commonly used F-score (Fsc) and intersection over union (IoU) metrics, and compute their mean values across the testing set, denoted as mFsc and mIoU, respectively. Moreover, the stochastic gradient descent with momentum (SGDM) \cite{lecun2015deep} is used to optimize the CNNs.

\begin{table*}[t]
    \caption{Performances of different AA-UNet variants on the Pothole-600 validation set, where (A) is the U-Net baseline; and (B)--(T) are different variants. Best Results are shown in bold type.}
    \centering
    \begin{tabular}{C{0.8cm}C{1.1cm}C{1.1cm}C{1.1cm}C{1.1cm}C{1.1cm}C{1.5cm}C{1.5cm}C{1.5cm}}
    \toprule
    \multicolumn{1}{c}{\multirow{2}{*}{No.}} & \multicolumn{5}{c}{Attention Aggregation Scheme} & \multicolumn{3}{c}{Evaluation Metrics} \\ \cmidrule(l){2-6} \cmidrule(l){7-9}
    \multicolumn{1}{c}{} & \multicolumn{1}{c}{1st} & \multicolumn{1}{c}{2nd} & \multicolumn{1}{c}{3rd} & \multicolumn{1}{c}{4th} & \multicolumn{1}{c}{5th} & \multicolumn{1}{c}{mFsc~($\%$)} & \multicolumn{1}{c}{mIoU~($\%$)} & \multicolumn{1}{c}{Runtime~(ms)} \\ \midrule
    (A) & -- & -- & -- & -- & -- & 75.9 & 61.2 & \textbf{31.3} \\ \midrule
    (B) & -- & -- & -- & -- & DAM & 79.7 & 66.3 & 33.7 \\
    (C) & -- & -- & -- & -- & CAM & 79.5 & 66.0 & 31.5 \\
    (D) & -- & -- & -- & -- & PAM & 79.4 & 65.9 & 31.7 \\
    (E) & -- & -- & -- & CAM & -- & 78.7 & 64.9 & 31.6 \\
    (F) & -- & -- & -- & PAM & -- & 78.5 & 64.6 & 31.9 \\
    (G) & -- & -- & CAM & -- & -- & 78.0 & 63.9 & 31.8 \\
    (H) & -- & -- & PAM & -- & -- & 77.7 & 63.5 & 32.0 \\
    (I) & -- & CAM & -- & -- & -- & 77.8 & 63.6 & 32.1 \\
    (J) & -- & PAM & -- & -- & -- & 77.5 & 63.2 & 32.6 \\
    (K) & CAM & -- & -- & -- & -- & 77.6 & 63.4 & 32.3 \\
    (L) & PAM & -- & -- & -- & -- & 77.8 & 63.7 & 33.5 \\ \midrule
    (M) & -- & -- & -- & CAM & DAM & 80.2 & 66.9 & 33.8 \\
    (N) & -- & -- & -- & PAM & DAM & 77.1 & 62.7 & 34.0 \\
    (O) & -- & -- & CAM & CAM & DAM & 80.7 & 67.6 & 33.9 \\
    (P) & -- & -- & PAM & CAM & DAM & 77.8 & 63.6 & 34.2 \\
    (Q) & -- & CAM & CAM & CAM & DAM & 81.0 & 68.0 & 34.1 \\
    (R) & -- & PAM & CAM & CAM & DAM & 79.7 & 66.2 & 34.5 \\
    (S) & CAM & CAM & CAM & CAM & DAM & 81.3 & 68.5 & 34.3 \\
    (T) & PAM & CAM & CAM & CAM & DAM & \textbf{82.6} & \textbf{70.3} & 34.7 \\
    \bottomrule
    \end{tabular}
    \label{tab.aa_unet}
\end{table*}

\begin{table*}[t]
    \caption{Performances of different AA-RTFNet variants on the Pothole-600 validation set, where (A) is the RTFNet baseline; and (B)--(J) are different variants. Best Results are shown in bold type.}
    \centering
    \begin{tabular}{C{0.8cm}C{1.1cm}C{1.1cm}C{1.1cm}C{1.1cm}C{1.1cm}C{1.5cm}C{1.5cm}C{1.5cm}}
    \toprule
    \multicolumn{1}{c}{\multirow{2}{*}{No.}} & \multicolumn{5}{c}{Attention Aggregation Scheme} & \multicolumn{3}{c}{Evaluation Metrics} \\ \cmidrule(l){2-6} \cmidrule(l){7-9}
    \multicolumn{1}{c}{} & \multicolumn{1}{c}{1st} & \multicolumn{1}{c}{2nd} & \multicolumn{1}{c}{3rd} & \multicolumn{1}{c}{4th} & \multicolumn{1}{c}{5th} & \multicolumn{1}{c}{mFsc~($\%$)} & \multicolumn{1}{c}{mIoU~($\%$)} & \multicolumn{1}{c}{Runtime~(ms)} \\ \midrule
    (A) & -- & -- & -- & -- & -- & 81.3 & 68.5 & \textbf{46.7} \\ \midrule
    (B) & -- & -- & -- & -- & DAM & 82.5 & 70.2 & 49.1 \\
    (C) & -- & -- & -- & CAM & DAM & 82.6 & 70.4 & 49.2 \\
    (D) & -- & -- & -- & PAM & DAM & 81.7 & 69.0 & 49.4 \\
    (E) & -- & -- & CAM & CAM & DAM & 82.8 & 70.7 & 49.3 \\
    (F) & -- & -- & PAM & CAM & DAM & 81.9 & 69.3 & 49.7 \\
    (G) & -- & CAM & CAM & CAM & DAM & 83.4 & 71.6 & 49.6 \\
    (H) & -- & PAM & CAM & CAM & DAM & 83.1 & 71.1 & 49.9 \\
    (I) & CAM & CAM & CAM & CAM & DAM & 84.1 & 72.5 & 50.0 \\
    (J) & PAM & CAM & CAM & CAM & DAM & \textbf{85.0} & \textbf{73.9} & 50.2 \\
    \bottomrule
    \end{tabular}
    \label{tab.aa_rtfnet}
\end{table*}

\subsection{Architecture Selection of AA-UNet and AA-RTFNet}
\label{sec.architecture_selection}
In this subsection, we conduct experiments to select the best architecture for our AA-UNet and AA-RTFNet. All the AA-UNet variants use the same training setups, so do all the AA-RTFNet variants. It should be noted here that $n=5$ is for both AA-UNet and AA-RTFNet. We also record the inference time of each variant on an NVIDIA GTX 1080Ti graphics card for comparison. (B)--(L) in Table~\ref{tab.aa_unet} present the effects of a single AM at different network levels. We can see that an AM can bring in better performance improvement when it is added at a higher level, as this can influence the subsequent processes. Moreover, DAM outperforms CAM and PAM at the highest level, since DAM adopts the self-attention scheme, which can achieve better performance, as mentioned above. Furthermore, our CAM performs better than our PAM at higher levels, since feature maps at higher levels have more channels but limited spatial sizes and it is more useful to apply weights on channels. Conversely, feature maps at lower levels have larger spatial sizes but limited channels, and thus it is more useful to adopt our PAM.

Based on these observations, we test the performance of different attention aggregation schemes for our AA-UNet and AA-RTFNet on the validation set, as shown on (M)--(T) in Table \ref{tab.aa_unet} and (B)--(J) in Table \ref{tab.aa_rtfnet}, respectively. We can see that adopting PAM at the lowest network level, adopting DAM at the highest network level, and adopting CAM at other network levels can achieve the best performance for both AA-UNet and AA-RTFNet. Compared with the baseline models, our AA-UNet and AA-RTFNet can increase the mIoU by 9.1\% and 5.4\%, respectively, with acceptable extra runtime, which demonstrates the effectiveness and efficiency of our attention aggregation framework.

\begin{figure}[!htbp]
    \centering
    \includegraphics[width=0.999\textwidth]{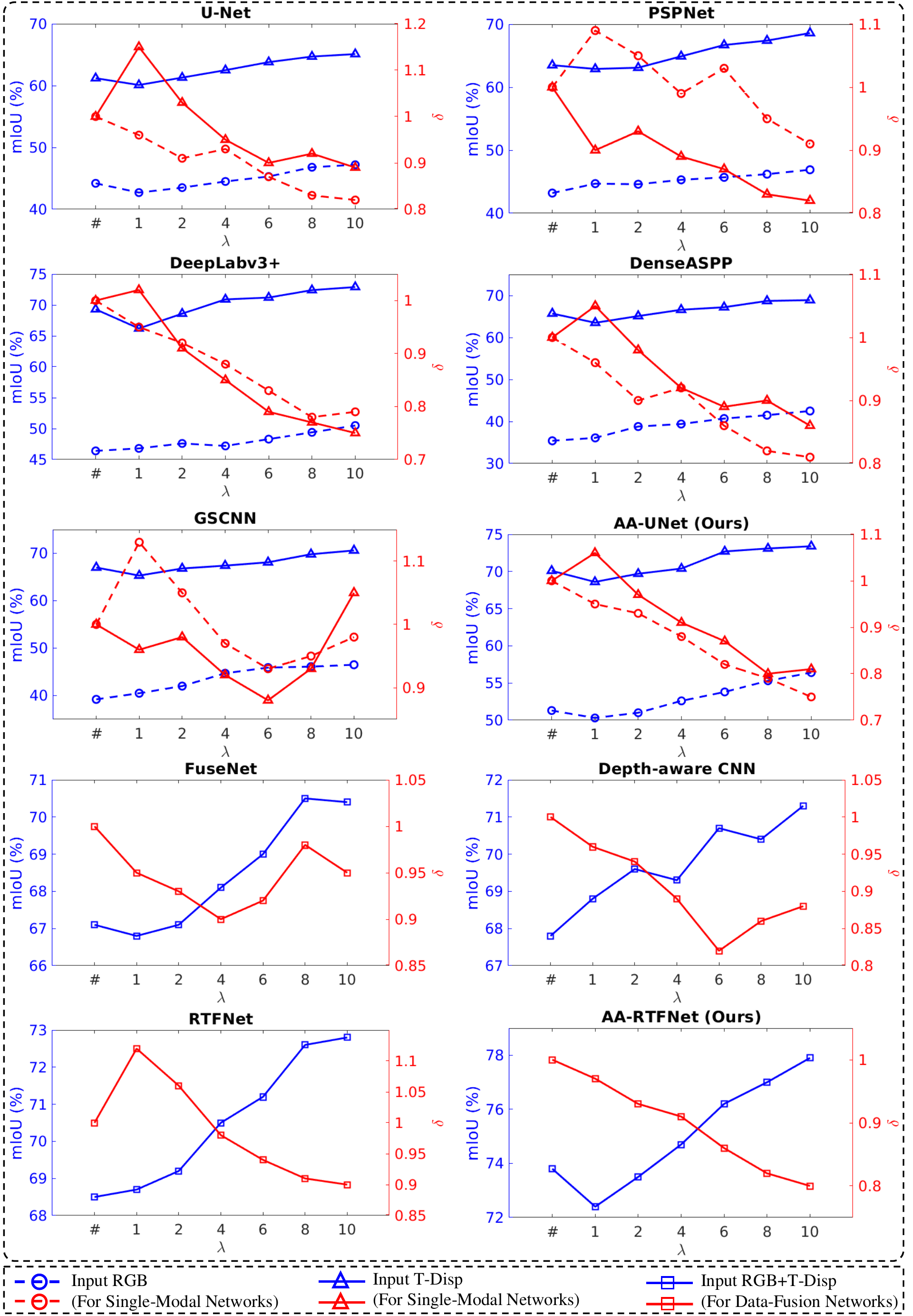}
    \caption{Performance comparison among eight SoA CNNs, AA-UNet and AA-RTFNet on the Pothole-600 testing set, where the symbol ``\#'' in the $\lambda$ axis  means that we use the original training set in the CNN. }
    \label{fig.evaluation}
\end{figure}

\begin{figure}[!htbp]
	\centering
	\includegraphics[width=0.999\textwidth]{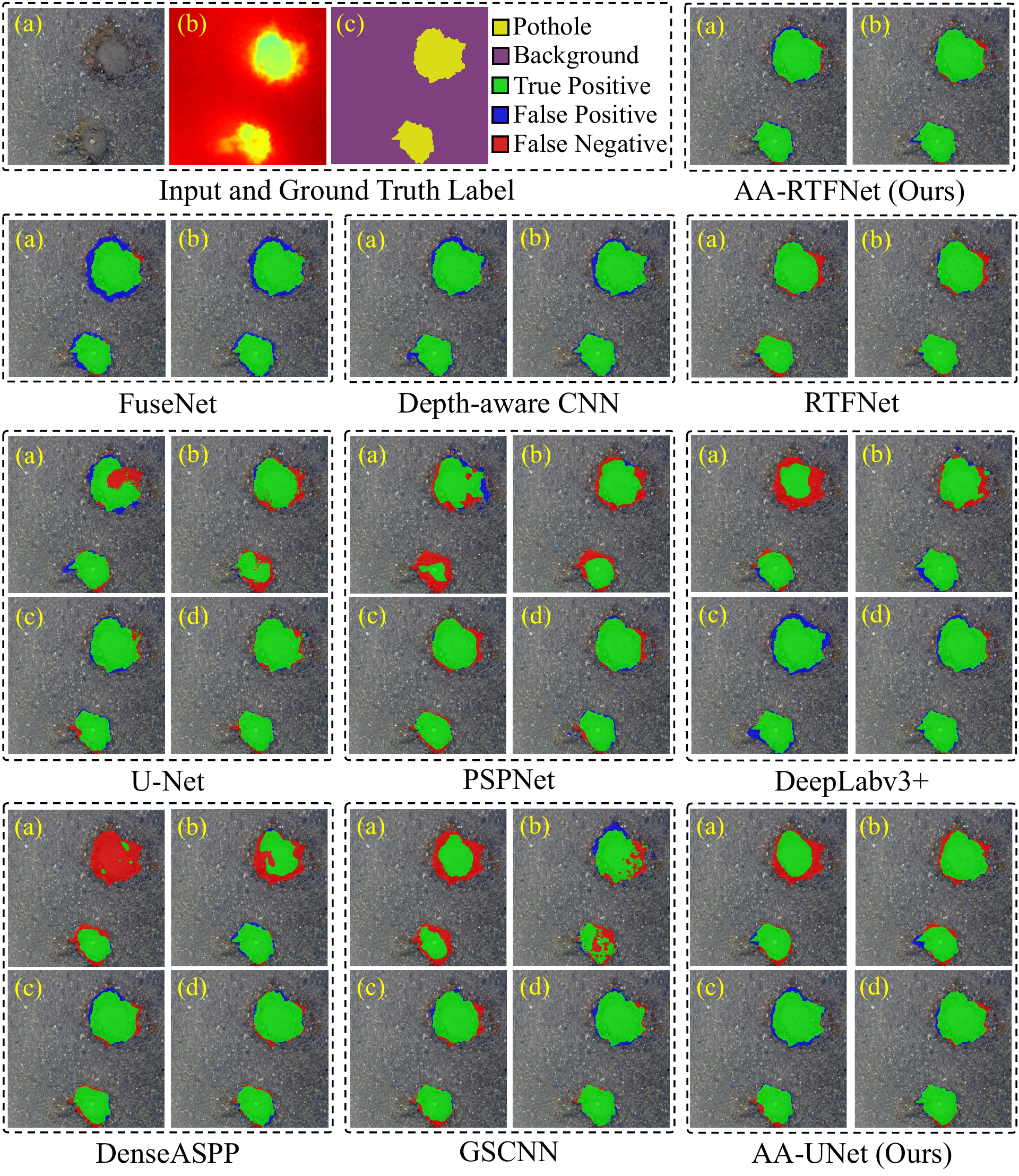}
	\caption{An example of the experimental results on the Pothole-600 testing set. For the input and ground truth label block: (a) RGB, (b) T-Disp, and (c) ground truth label; For the single-modal network (including U-Net \cite{ronneberger2015u}, PSPNet \cite{zhao2017pyramid}, DeepLabv3+ \cite{chen2018encoder}, DenseASPP \cite{yang2018denseaspp}, GSCNN \cite{takikawa2019gated} and our AA-UNet) blocks: (a) input RGB from the original training set, (b) input RGB from the whole augmented training set, (c) input T-Disp from the original training set, and (d) input T-Disp from the whole augmented training set; For the data-fusion network (including FuseNet \cite{hazirbas2016fusenet}, Depth-aware CNN \cite{wang2018depth}, RTFNet \cite{sun2019rtfnet} and our AA-RTFNet) blocks: (a) input RGB+T-Disp from the original training set, and (b) input RGB+T-Disp from the whole augmented training set.}
	\label{fig.comparison}
\end{figure}

\subsection{Performance Evaluation of Road Pothole Detection}
\label{sec.performance_evaluation}
In this subsection, we evaluate the performance of our AA-UNet and AA-RTFNet both qualitatively and quantitatively on the testing set. As mentioned previously, we use different numbers of images selected from the augmented training set to train each CNN.  $\lambda$ denotes the number of samples used in the augmented training set versus the number of samples in the original training set. For example, $\lambda=2$ means that we train the CNN with $240 \times 2 = 480$ samples randomly selected from the augmented training set. In addition, we introduce a new evaluation metric $\delta$ for better comparison. For a given training setup, $\delta$ is defined as ratio of the number of iterations for the network to converge using the augmented training set to that of the original training set. $\delta < 1$ means that the training setup converges faster than the baseline setup.

The quantitative results are shown in Fig. \ref{fig.evaluation}, where we can clearly observe that the single-modal CNNs with our transformed disparity images as inputs generally perform better than they do with RGB images as inputs, and the mIoU increases by about 17--31\%. This is because our transformed disparity images can make the road potholes become highly distinguishable, and can thus benefit all CNNs for road pothole detection. Moreover, we can see that when $\lambda \geq 4$, the CNNs trained with the augmented training set generally outperform themselves when trained with the original training set, and $\delta < 1$ holds in most cases, which demonstrates that adversarial domain adaptation can not only significantly improve pothole detection accuracy but  can also accelerate the network convergence. Compared with the training setup using the original training set, an increase of around 3--8\% is witnessed on the mIoU for the training setup using the whole augmented training set. This is because these two sets share very similar distributions, and our augmented training set possesses an expanded distribution, which can improve road pothole detection performance. In addition, our AA-UNet and AA-RTFNet outperform all other SoA single-modal and data-fusion networks for road pothole detection, respectively, which strongly validates the effectiveness and efficiency of our attention aggregation framework. Readers can see that our AA-UNet can increase the mIoU by approximately 3--14\% compared with the SoA single-modal networks, and our AA-RTFNet can increase the mIoU by about 5--8\% compared with the SoA data-fusion networks. The qualitative results shown in Fig. \ref{fig.comparison} can also confirm the superiority of our proposed approaches.

\section{Conclusion}
\label{sec.conclusion}
The major contributions of this paper include: a) a novel attention aggregation framework, which can help the CNNs focus more on salient objects, such as road potholes, so as to improve semantic segmentation for better  pothole detection results; b) a novel training set augmentation technique developed based on adversarial domain adaptation, which can produce more synthetic road RGB images and their corresponding transformed road disparity (or inverse depth) images to improve both the efficiency and accuracy of CNN training; c) a large-scale road pothole detection dataset, publicly available at \url{sites.google.com/view/pothole-600} for research purposes. The experimental results validated the effectiveness and feasibility of our proposed attention aggregation framework and the training set augmentation technique for enhancing road pothole detection. Moreover, we believe our proposed techniques can also be used for many other semantic segmentation applications, such as freespace detection.

\section*{Acknowledgements}
This work was supported by the National Natural Science Foundation of China, under grant No. U1713211, Collaborative Research Fund by Research Grants Council Hong Kong, under Project No. C4063-18G, and the Research Grant Council of Hong Kong SAR Government, China, under Project No. 11210017, awarded to Prof. Ming Liu.

\clearpage


\begin{thebibliography}{10}
	
	\bibitem{mathavan2015review}
	Mathavan, S., Kamal, K., Rahman, M.:
	\newblock A review of three-dimensional imaging technologies for pavement
	distress detection and measurements.
	\newblock IEEE Transactions on Intelligent Transportation Systems
	\textbf{16}(5) (2015)  2353--2362
	
	\bibitem{fan2019pothole}
	Fan, R., Ozgunalp, U., Hosking, B., Liu, M., Pitas, I.:
	\newblock Pothole detection based on disparity transformation and road surface
	modeling.
	\newblock IEEE Transactions on Image Processing \textbf{29} (2019)  897--908
	
	\bibitem{koch2015review}
	Koch, C., Georgieva, K., Kasireddy, V., Akinci, B., Fieguth, P.:
	\newblock A review on computer vision based defect detection and condition
	assessment of concrete and asphalt civil infrastructure.
	\newblock Advanced Engineering Informatics \textbf{29}(2) (2015)  196--210
	
	\bibitem{pothole_news_dani_king}
	Majendie, M.:
	\newblock Dani king: 'it was just a freak accident but i thought i was going to
	die'.
	\newblock Technical report, Independent (June 2015)
	
	\bibitem{fan2019road}
	Fan, R., Liu, M.:
	\newblock Road damage detection based on unsupervised disparity map
	segmentation.
	\newblock IEEE Transactions on Intelligent Transportation Systems (2019)
	
	\bibitem{sandiego_pothole}
	Devine, R.:
	\newblock City of san diego asking residents to report potholes.
	\newblock Technical report, NBC San Diego (January 2017)
	
	\bibitem{bbc_pothole_repairing}
	News, B.:
	\newblock Government to pledge billions for filling potholes.
	\newblock Technical report, BBC News (March 2020)
	
	\bibitem{fan2019real}
	Fan, R., Jiao, J., Pan, J., Huang, H., Shen, S., Liu, M.:
	\newblock Real-time dense stereo embedded in a uav for road inspection.
	\newblock In: Proceedings of the IEEE Conference on Computer Vision and Pattern
	Recognition (CVPR) Workshops, IEEE (2019)  535--543
	
	\bibitem{leo2018deep}
	Leo, M., Furnari, A., Medioni, G.G., Trivedi, M., Farinella, G.M.:
	\newblock Deep learning for assistive computer vision.
	\newblock In: Proceedings of the European Conference on Computer Vision (ECCV).
	(2018)  0--0
	
	\bibitem{land_rover_pothole}
	Rover, J.L.:
	\newblock Pothole detection technology research announced by jaguar land rover
	
	\bibitem{ford_pothole}
	Baraniuk, C.:
	\newblock Ford developing pothole alert system for drivers (February 2017)
	
	\bibitem{jahanshahi2013unsupervised}
	Jahanshahi, M.R., Jazizadeh, F., Masri, S.F., Becerik-Gerber, B.:
	\newblock Unsupervised approach for autonomous pavement-defect detection and
	quantification using an inexpensive depth sensor.
	\newblock Journal of Computing in Civil Engineering \textbf{27}(6) (2013)
	743--754
	
	\bibitem{otsu1979threshold}
	Otsu, N.:
	\newblock A threshold selection method from gray-level histograms.
	\newblock IEEE transactions on systems, man, and cybernetics \textbf{9}(1)
	(1979)  62--66
	
	\bibitem{long2015fully}
	Long, J., Shelhamer, E., Darrell, T.:
	\newblock Fully convolutional networks for semantic segmentation.
	\newblock In: Proceedings of the IEEE Conference on Computer Vision and Pattern
	Recognition. (2015)  3431--3440
	
	\bibitem{ronneberger2015u}
	Ronneberger, O., Fischer, P., Brox, T.:
	\newblock U-net: Convolutional networks for biomedical image segmentation.
	\newblock In: International Conference on Medical Image Computing and
	Computer-assisted Intervention, Springer (2015)  234--241
	
	\bibitem{zhao2017pyramid}
	Zhao, H., Shi, J., Qi, X., Wang, X., Jia, J.:
	\newblock Pyramid scene parsing network.
	\newblock In: Proceedings of the IEEE Conference on Computer Vision and Pattern
	Recognition. (2017)  2881--2890
	
	\bibitem{chen2018encoder}
	Chen, L.C., Zhu, Y., Papandreou, G., Schroff, F., Adam, H.:
	\newblock Encoder-decoder with atrous separable convolution for semantic image
	segmentation.
	\newblock In: Proceedings of the European Conference on Computer Vision (ECCV).
	(2018)  801--818
	
	\bibitem{yang2018denseaspp}
	Yang, M., Yu, K., Zhang, C., Li, Z., Yang, K.:
	\newblock Denseaspp for semantic segmentation in street scenes.
	\newblock In: Proceedings of the IEEE Conference on Computer Vision and Pattern
	Recognition. (2018)  3684--3692
	
	\bibitem{takikawa2019gated}
	Takikawa, T., Acuna, D., Jampani, V., Fidler, S.:
	\newblock Gated-scnn: Gated shape cnns for semantic segmentation.
	\newblock In: Proceedings of the IEEE International Conference on Computer
	Vision. (2019)  5229--5238
	
	\bibitem{wang2020applying}
	Wang, H., Fan, R., Sun, Y., Liu, M.:
	\newblock Applying surface normal information in drivable area and road anomaly
	detection for ground mobile robots.
	\newblock In: 2020 IEEE/RSJ International Conference on Intelligent Robots and
	Systems (IROS). (2020) to be published.
	
	\bibitem{fan2020sne-roadseg}
	Fan, R., Wang, H., Cai, P., Liu, M.:
	\newblock Sne-roadseg: Incorporating surface normal information into semantic
	segmentation for accurate freespace detection.
	\newblock In: European Conference on Computer Vision (ECCV), Springer (2020)
	
	\bibitem{hazirbas2016fusenet}
	Hazirbas, C., Ma, L., Domokos, C., Cremers, D.:
	\newblock Fusenet: Incorporating depth into semantic segmentation via
	fusion-based cnn architecture.
	\newblock In: Asian Conference on Computer Vision, Springer (2016)  213--228
	
	\bibitem{wang2018depth}
	Wang, W., Neumann, U.:
	\newblock Depth-aware cnn for rgb-d segmentation.
	\newblock In: Proceedings of the European Conference on Computer Vision (ECCV).
	(2018)  135--150
	
	\bibitem{sun2019rtfnet}
	Sun, Y., Zuo, W., Liu, M.:
	\newblock Rtfnet: Rgb-thermal fusion network for semantic segmentation of urban
	scenes.
	\newblock IEEE Robotics and Automation Letters \textbf{4}(3) (2019)  2576--2583
	
	\bibitem{vaswani2017attention}
	Vaswani, A., Shazeer, N., Parmar, N., Uszkoreit, J., Jones, L., Gomez, A.N.,
	Kaiser, {\L}., Polosukhin, I.:
	\newblock Attention is all you need.
	\newblock In: Advances in neural information processing systems. (2017)
	5998--6008
	
	\bibitem{hu2018squeeze}
	Hu, J., Shen, L., Sun, G.:
	\newblock Squeeze-and-excitation networks.
	\newblock In: Proceedings of the IEEE conference on computer vision and pattern
	recognition. (2018)  7132--7141
	
	\bibitem{wang2018non}
	Wang, X., Girshick, R., Gupta, A., He, K.:
	\newblock Non-local neural networks.
	\newblock In: Proceedings of the IEEE conference on computer vision and pattern
	recognition. (2018)  7794--7803
	
	\bibitem{yuan2018ocnet}
	Yuan, Y., Wang, J.:
	\newblock Ocnet: Object context network for scene parsing.
	\newblock arXiv preprint arXiv:1809.00916 (2018)
	
	\bibitem{fu2019dual}
	Fu, J., Liu, J., Tian, H., Li, Y., Bao, Y., Fang, Z., Lu, H.:
	\newblock Dual attention network for scene segmentation.
	\newblock In: Proceedings of the IEEE Conference on Computer Vision and Pattern
	Recognition. (2019)  3146--3154
	
	\bibitem{huang2019ccnet}
	Huang, Z., Wang, X., Huang, L., Huang, C., Wei, Y., Liu, W.:
	\newblock Ccnet: Criss-cross attention for semantic segmentation.
	\newblock In: Proceedings of the IEEE International Conference on Computer
	Vision. (2019)  603--612
	
	\bibitem{goodfellow2014generative}
	Goodfellow, I., Pouget-Abadie, J., Mirza, M., Xu, B., Warde-Farley, D., Ozair,
	S., Courville, A., Bengio, Y.:
	\newblock Generative adversarial nets.
	\newblock In: Advances in neural information processing systems. (2014)
	2672--2680
	
	\bibitem{zhu2017unpaired}
	Zhu, J.Y., Park, T., Isola, P., Efros, A.A.:
	\newblock Unpaired image-to-image translation using cycle-consistent
	adversarial networks.
	\newblock In: Proceedings of the IEEE international conference on computer
	vision. (2017)  2223--2232
	
	\bibitem{isola2017image}
	Isola, P., Zhu, J.Y., Zhou, T., Efros, A.A.:
	\newblock Image-to-image translation with conditional adversarial networks.
	\newblock In: Proceedings of the IEEE conference on computer vision and pattern
	recognition. (2017)  1125--1134
	
	\bibitem{fan2018road}
	Fan, R., Ai, X., Dahnoun, N.:
	\newblock Road surface 3d reconstruction based on dense subpixel disparity map
	estimation.
	\newblock IEEE Transactions on Image Processing \textbf{27}(6) (2018)
	3025--3035
	
	\bibitem{labayrade2003single}
	Labayrade, R., Aubert, D.:
	\newblock A single framework for vehicle roll, pitch, yaw estimation and
	obstacles detection by stereovision.
	\newblock In: IEEE IV2003 Intelligent Vehicles Symposium. Proceedings (Cat. No.
	03TH8683), IEEE (2003)  31--36
	
	\bibitem{lecun2015deep}
	LeCun, Y., Bengio, Y., Hinton, G.:
	\newblock Deep learning.
	\newblock nature \textbf{521}(7553) (2015)  436--444
	
\end{thebibliography}

\end{document}